%% file: main.tex
\documentclass[10pt,twocolumn,letterpaper]{article}

\usepackage{cvpr}              
\usepackage[accsupp]{axessibility} 
\usepackage{graphicx}
\graphicspath{ {./images/} }

\usepackage{multirow}
\input{preamble}

\definecolor{cvprblue}{rgb}{0.21,0.49,0.74}
\usepackage[pagebackref,breaklinks,colorlinks,citecolor=cvprblue]{hyperref}

\def\ours{\texttt{\textbf{SViTT-Ego}}}

\def\paperID{15} 
\def\confName{CVPR}
\def\confYear{2024}

\title{SViTT-Ego: A Sparse Video-Text Transformer for Egocentric Video}

\author{Hector A. Valdez\\
Intel Labs\\
{\tt\small hector.a.valdez@intel.com}
\and
Kyle Min\\
Intel Labs\\
{\tt\small kyle.min@intel.com}
\and
Subarna Tripathi\\
Intel Labs\\
{\tt\small subarna.tripathi@intel.com}
}

\begin{document}
\maketitle
\input{sec/0_abstract}    
\input{sec/1_intro}
\input{sec/2_related}
\input{sec/3_method}
\input{sec/4_results}
\input{sec/5_conclusion}

{
    \small
    \bibliographystyle{ieeenat_fullname}
    \bibliography{main}
}


\end{document}

%% file: preamble.tex
\usepackage[dvipsnames]{xcolor}


%% file: sec/0_abstract.tex
\begin{abstract}
Pretraining egocentric vision-language models has become essential to improving downstream egocentric video-text tasks. These egocentric foundation models commonly use the transformer architecture. The memory footprint of these models during pretraining can be substantial. Therefore, we pretrain \ours, the first sparse egocentric video-text transformer model integrating edge and node sparsification. We pretrain on the EgoClip dataset and incorporate the egocentric-friendly objective EgoNCE, instead of the frequently used InfoNCE. Most notably, \ours\ obtains a +$2.8\%$ gain on EgoMCQ (intra-video) accuracy compared to \textsc{L}\textsc{a}\textsc{V}\textsc{i}\textsc{L}\textsc{a}\textsubscript{\textsc{large}}, with no additional data augmentation techniques other than standard image augmentations, yet pretrainable on memory-limited  devices.
\end{abstract}

%% file: sec/1_intro.tex
\section{Introduction}
\label{sec:intro}

Pretraining vision-language models is a vital step in constructing foundation models capable of efficient fine-tuning across numerous downstream tasks or, more importantly, zero-shot scenarios. Recently, there has been a surge of interest in egocentric videos prompting a variety of works \citep{lin2022egocentric, pramanick2023egovlpv2, ashutosh2023hiervl, zhao2022learning} aimed at pretraining video-text models tailored for egocentric downstream applications. The prevailing model architecture that is used for both vision and text encoder is the transformer. 

At the heart of the transformer lies the attention mechanism having quadratic time and space complexity with respect to its input. Memory bottlenecks can quickly arise, particularly when handling multi-frame image inputs in the vision encoder. To the best of our knowledge, we are the first to address this issue in the egocentric vision-language paradigm. We propose \ours, and approach the memory bottleneck problem by applying edge and node sparsification to the memory-hungry video and cross-modal encoders.

The key contributions of this work are: 
(1) a video-text architecture \ours\ that uses edge and node sparsity for egocentric videos; 
(2) empirical results validating EgoNCE as a superior objective for intra-video EgoMCQ scenarios compared to the InfoNCE objective; 
(3) state-of-the-art performance on intra-video EgoMCQ task; and
(4) efficient pretraining on memory-constrained devices. 

%% file: sec/2_related.tex
\section{Related Work}
\label{sec:related_work}

\textbf{Egocentric video-language pretraining.} 
\noindent Vision-language pretraining has seen significant adoption for a number of egocentric video-text downstream tasks. EgoVLP \citep{lin2022egocentric} is one of the first egocentric vision-language models that is pretrained on the Ego4D \citep{grauman2022ego4d} dataset and proposes the EgoNCE objective to optimize model parameters. EgoVLPv2 \citep{pramanick2023egovlpv2} is a continuation of the EgoVLP work which adds cross-modal fusion to both the vision and text backbones. HierVL \citep{ashutosh2023hiervl} uses a hierarchical video-language model that captures both short-term actions and long-term intents in the video. L\textsc{a}V\textsc{i}L\textsc{a} \citep{zhao2022learning} uses a Large Language Model (LLM) to rephrase Ego4D's annotations and synthetically generate more text descriptions per video sample, which augmented their training set by $15\times$. In our work, we use the EgoNCE objective for egocentric-friendly pretraining.
\newline

\noindent \textbf{Sparse transformers.} 
The transformer's attention mechanism has quadratic time and space complexity. Recently, several token sparsification methods have been proposed to improve the efficiency of vision transformers. For instance, DynamicViT~\cite{rao2021dynamicvit} and EViT~\cite{liang2022not} reduce the number of input tokens by identifying the less informative ones. ToMe~\cite{bolya2022token} shows that gradually combining similar tokens significantly improves the throughput of transformers. SViTT \citep{svitt2023} uses edge sparsity to reduce query-key pairs while maintaining its global reasoning capability and uses node sparsity to minimize the usage of non-informative visual tokens. We adopt SViTT for its highly efficient training and inference schemes in this work.

%% file: sec/3_method.tex
\section{Methodology}

\noindent \textbf{EgoClip.} We pretrain our model on the EgoClip dataset \citep{lin2022egocentric}, which consists of 3.8 million clip-text pairs selected from Ego4D \citep{grauman2022ego4d}. The videos cover a diverse range of human daily activities, with a duration of 2.9K video hours. Each clip-text pair consists of a 1-second clip (30 frames per second) and a text description of actions occurring within the clip.
\newline

\noindent \textbf{Sampling Strategies.} We adopt the same sampling strategies: action-aware positive sampling and scene-aware negative sampling. Action-aware positive samples are text descriptions that share at least one noun and one verb, where nouns and verbs can be synonyms based on the Ego4D dictionary taxonomy. Scene-aware negative samples are considered hard negatives since they are different actions occurring in the same video scene as the positive sample. These hard negatives are temporally adjacent to their positive counterparts and do not overlap in frames.
\newline

\noindent \textbf{EgoNCE.} Given a batch $\mathcal{B}$, let the positive sample set be $P_{i} = \{ j \in \mathcal{B} \mid \text{noun}(j) \cap \text{noun}(i) \not = \emptyset, \text{verb}(j) \cap \text{verb}(i) \not = \emptyset \}$. The original batch is augmented with hard negative samples to create an updated batch $\tilde{\mathcal{B}}$. We optimize the duel encoder using the EgoNCE objective \citep{lin2022egocentric}:

\newcommand{\numerator}{\sum_{k \in P_{i}}{\exp{(\textbf{v}_{i}^{T} \textbf{t}_{k} \slash \tau)}}}
\newcommand{\denominator}{\sum_{j \in \mathcal{B}} (\exp{(\textbf{v}_{i}^{T} \textbf{t}_{j} \slash \tau)} + \exp{(\textbf{v}_{i}^{T} \textbf{t}_{j'} \slash \tau)})}

\begin{equation}
L = \frac{1}{ |\tilde{\mathcal{B}}| } \sum_{i \in \tilde{\mathcal{B}}}{ \log{ \frac{ \numerator }{ \denominator } } }
\end{equation}

\noindent where $\textbf{v}$ is the [CLS] video feature vector and $\textbf{t}$ is the [CLS] text feature vector.

%% file: sec/4_results.tex

\section{Results}
\label{sec:results}

\subsection{Implementation Details}\label{implement_details}
We use the same sparse frame sampling strategy as in \citep{svitt2023} and sample $4$ frames per video. Each frame is randomly cropped, and then resized into a spatial resolution of $224 \times 224$. Then, each image is decomposed into $14 \times 14$ spatial patches. We use an AdamW optimizer \citep{loshchilov2019decoupled} with an initial learning rate of $3 \times 10^{-5}$ ($\beta_{1} = 0.9, \beta_{2} = 0.999$) and a weight decay of $0.02$, with a cosine learning rate schedule and warm-up of one epoch over $10$ epochs. We use PyTorch's native FP16 mixed precision and distributed training with a per GPU batch size of 8 over 8 GPUs, each GPU with 12 GB of memory. 
\newline

%

%

\noindent \textbf{Architecture.} Our implementation of sparse video-text transformer is based on SViTT \citep{svitt2023}, which is composed of a video encoder, text encoder, and multimodal encoder. The video and text encoder can be used as a duel encoder for specific vision-language downstream tasks, while the multimodal encoder fuses video features with text features using cross-attention for other downstream tasks. Similar to SViTT, the video encoder is a 12-layer BEiT-B \citep{bao2022beit} initialized with ImageNet weights and inflated for video inputs. The text encoder is initialized with BERT\textsubscript{BASE} \citep{devlin2019bert} weights, whose last 3 layers are modified to implement the multimodal encoder.
\newline

\noindent \textbf{Sparsification.}
Ablation studies in \citep{svitt2023} showed the best balance between sparsity and performance was using both edge and node sparsification. We set two different edge sparisty configurations $(K_{l}, K_{r}, G) = (1, 3, 56)$ and $(1, 5, 56)$, where $K_{l}$ is sparse local attention, $K_{r}$ is sparse random attention, and $G$ is attention block size. We set node sparsity to $(q_{v}, q_{m}) = (0.7, 0.1)$ where $q_{v}$ is vision token keep rate and $q_{m}$ is multimodal token keep rate. The token keep strategy uses [CLS] tokens.
\newline

\noindent \textbf{Pretraining.} \ours\ is pretrained on 3.8M video-text pairs from EgoClip. We optimize the duel encoder by projecting the [CLS] tokens of the video and text encoders into a joint representation space where we use the contrastive EgoNCE loss \citep{lin2022egocentric}. The output of the multimodal encoder is optimized with video-text matching (VTM) and masked language modeling (MLM) losses, same as \citep{svitt2023}.
\newline

\noindent \textbf{EgoMCQ.} Egocentric Multiple Choice Question task contains 39,751 questions created from Ego4D. The task requires selecting the correct video clip from five choices given a text query description. 
\newline

\noindent \textbf{EgoNLQ.} Ego4D Natural Language Queries task is also constructed from Ego4D. The task requires localizing the temporal window within a video, given a natural language question.
\newline

\subsection{Initial Experiment}
Before we performed full pretraining on the entire 3.8M EgoClip dataset, we conducted a quick experiment to evaluate duel encoder objectives InfoNCE \citep{oord2019representation} and EgoNCE. We use the same configuration as mentioned in \ref{implement_details}, except we fix edge sparisty to $(K_{l}, K_{r}, G) = (1, 3, 56)$ and use a random $10\%$ subset of the 3.8M dataset for each epoch.

When using EgoMCQ as a metric, results in Tab. \ref{table:initial_results} confirm findings in \citep{lin2022egocentric} showing EgoNCE as a better objective to optimize model parameters compared to the naive InfoNCE. We use 4 frames during inference to gather \ours\ performance results.

\begin{table}[h!]
\centering
\begin{tabular}{lccc}
\hline
                & \textbf{\# Pretrain}  & \multicolumn{2}{c}{\textbf{EgoMCQ}} \\ 
\textbf{Objective} & \textbf{Dataset}    & \multicolumn{2}{c}{Accuracy (\%)} \\ 
                &      (per epoch)       & Inter         & Intra \\ \hline
InfoNCE & 10\%  & 89.0 & 43.3 \\ 
EgoNCE & 10\%  & \textbf{89.4} & \textbf{53.0} \\ 
\end{tabular}
\caption{\label{table:initial_results} In our initial experiment, EgoNCE performs significantly better on intra-video scenarios compared to InfoNCE by a margin of +$9.7\%$ accuracy.}
\end{table}

\subsection{Pretraining Result}
We compare pretrained \ours\ to state-of-the-art vision-language models on the validation set of Egocentric Multiple Choice Question (EgoMCQ). Results in Tab. \ref{table:ValidPerf} show both \ours\ configurations $(K_{l}, K_{r}, G) = (1, 3, 56)$ and $(1, 5, 56)$ outperform EgoVLP and EgoVLPv2, when all models used the same amount of training data. Moreover, \ours\ exceeds L\textsc{a}V\textsc{i}L\textsc{a}\textsubscript{\textsc{base}} and L\textsc{a}V\textsc{i}L\textsc{a}\textsubscript{\textsc{large}} Intra-video accuracy even though both L\textsc{a}V\textsc{i}L\textsc{a} models are pretrained on 15× more narrations generated by GPT-2 \citep{radford2019gpt2}. We use 4 frames during inference to gather \ours\ performance results.
\newline

\begin{table}[h!]
\centering
\begin{tabular}{lccc}
\hline
                & \textbf{\# Pretrain}  & \multicolumn{2}{c}{\textbf{EgoMCQ}} \\ 
\textbf{Method} & \textbf{Dataset}      & \multicolumn{2}{c}{Accuracy (\%)} \\ 
                &                       & Inter         & Intra \\ \hline
L\textsc{a}V\textsc{i}L\textsc{a}\textsubscript{\textsc{base}} \citep{zhao2022learning}  & 56M & 93.8   & 59.9 \\ 
L\textsc{a}V\textsc{i}L\textsc{a}\textsubscript{\textsc{large}} \citep{zhao2022learning} & 56M & 94.5 & 63.1 \\ 
HierVL-Avg \citep{ashutosh2023hiervl} & 3.8M  & 90.3  & 53.1 \\ 
HierVL-SA \citep{ashutosh2023hiervl}  & 3.8M  & 90.5  & 52.4 \\ 
EgoVLP \citep{lin2022egocentric}                        & 3.8M  & 90.6          & 57.2 \\ 
EgoVLPv2 \citep{pramanick2023egovlpv2}                  & 3.8M  & 91.0          & 60.9 \\ \hline
\textbf{SViTT-Ego}\textsubscript{$K_{r}=3$} & 3.8M  & 92.9          & \textbf{65.1} \\ 
\textbf{SViTT-Ego}\textsubscript{$K_{r}=5$} & 3.8M  & 92.9          & \textbf{65.9} \\ 
\end{tabular}
\caption{\label{table:ValidPerf} \ours\ outperforms all state-of-the-art models on intra-video accuracy. When considering models trained solely on 3.8M samples without narration augmentations, \ours\ outperforms all models in inter-video and intra-video accuracy.}
\end{table}

\subsection{SViTT-Ego Video Features for EgoNLQ}
At the time of our work, GroundNLQ \citep{hou2023groundnlq} was 1st place on the Ego4D NLQ leaderboard so we used it as our video grounding model. However, there are some differences between our use of GroundNLQ and the original work due to computational constraints. The differences can be viewed in Tab. \ref{table:GroundingModelDiff}. We use \ours\ configuration $(K_{l}, K_{r}, G) = (1, 3, 56)$ to extract video features.
\newline

\begin{table}[h!]
\centering
\small\addtolength{\tabcolsep}{-3pt}
\scriptsize
\begin{tabular}{lllccc}
\hline 
 & \textbf{Video} & \textbf{Ego4D} & \textbf{Frame} &  & \textbf{Hidden}\\
\textbf{Method} & \textbf{Encoder} & \textbf{Dataset} & \textbf{Window} & \textbf{Stride} & \textbf{Dim} \\
\hline
GroundNLQ & InternVideo & NLQ, MQ, VQ & 16 & 16 & 384 \\
Ours & SViTT-Ego\textsubscript{$K_{r}=3$} & NLQ, MQ & 4 & 4 & 128 \\
\end{tabular}
\caption{\label{table:GroundingModelDiff} Differences between original GroundNLQ and our implementation.}
\end{table}

\noindent Pretraining and finetuning GroundNLQ using \ours\ features shows competitive IoU=0.3 and IoU=0.5 performance in Tab. \ref{table:NLQTestPerf}. All grounding models that were pretrained have better performance than those that were only finetuned. Among the pretrained models, our implementation under performs when compared to the original GroundNLQ and NaQ++ (ReLER + NaQ). We speculate that our implementation was at a disadvantage since it has less model parameters and did not include EgoVQ samples in the pretraining stage. 
\newline

\begin{table}[h!]
\centering
\small\addtolength{\tabcolsep}{-4pt}
\scriptsize
\begin{tabular}{llccccc}
\hline 
\textbf{Video} & \textbf{Grounding} & & \multicolumn{2}{c}{\textbf{IoU=0.3}} & \multicolumn{2}{c}{\textbf{IoU=0.5}} \\ 
\textbf{Encoder} & \textbf{Model} & \textbf{PT} & R@1 & R@5 & R@1 & R@5 \\ \hline
EgoVLP \citep{lin2022egocentric} & VSLNet \citep{zhang2020spanbased} & no & 10.46          & 16.76 & 6.24 & 11.29\\ 
SlowFast \citep{feichtenhofer2019slowfast} &&& \\ 
+ Omnivore \citep{girdhar2022omnivore} & ActionFormer \citep{zhang2022actionformer} & no & 15.71 & 28.45 & 9.57 & 18.03\\
+ EgoVLP \citep{lin2022egocentric} &&& \\
InternVideo \citep{wang2022internvideo} & VSLNet \citep{zhang2020spanbased} & no & 16.46 & 22.95 & 10.06 & 16.11 \\
InternVideo \citep{wang2022internvideo} & ReLER \citep{liu2022relerzjualibaba} + NaQ \citep{ramakrishnan2023naq} & yes & 21.70 & 25.12 & 13.64 & 16.33 \\ 
InternVideo \citep{wang2022internvideo} & GroundNLQ \citep{hou2023groundnlq} & yes & 25.67 & 42.06 & 18.18 & 29.80\\ \hline
\textbf{SViTT-Ego}\textsubscript{$K_{r}=3$} & GroundNLQ \citep{hou2023groundnlq} & yes & 19.41 & 33.52 & 13.29 & 24.28\\ 
\end{tabular}
\caption{\label{table:NLQTestPerf} Recall for IoU=0.3 and IoU=0.5 on Ego4D NLQ challenge test set.}
\end{table}

\subsection{Other Remarks}

\begin{figure}[h]
\title{\textbf{EgoClip annotation:} puts the speaker and notepad in both hands on a seat}
\centering
\includegraphics[width=8cm]{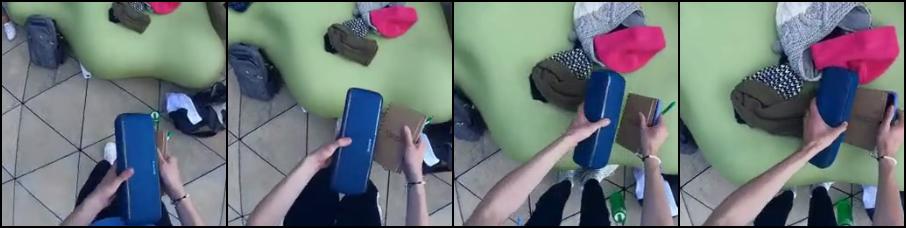}
\includegraphics[width=8cm]{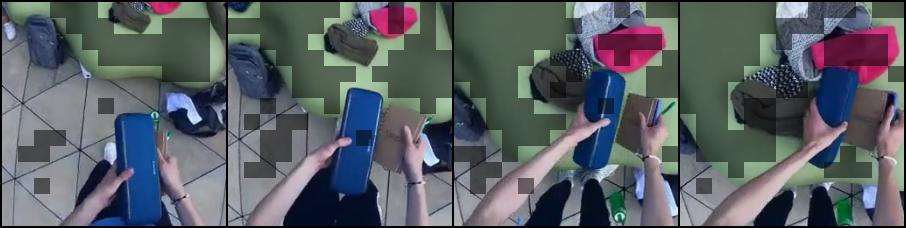}
\includegraphics[width=8cm]{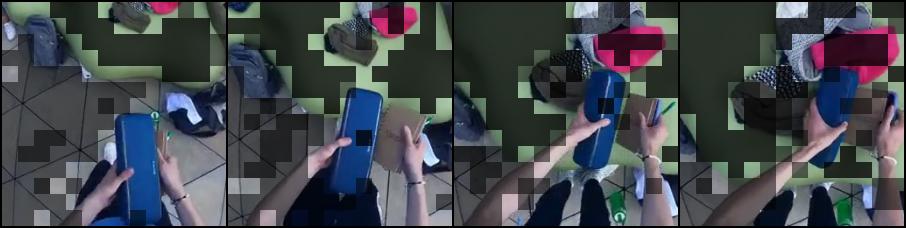}
\includegraphics[width=8cm]{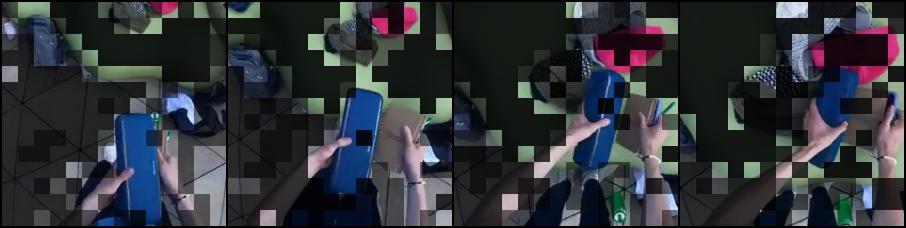}
\centering
\caption{Given $q_{v}=0.7$, we show the following qualitative results with the vision encoder:  
row 1, shows 4 frame input;  
row 2, shows video encoder's layer 4 after visual token pruning;  
row 3, shows video encoder's layer 7 after visual token pruning; and
row 4, shows video encoder's layer 10 after visual token pruning. We follow \citep{liang2022patches} to prune visual tokens.}
\label{figure:visualize_sparsification}
\end{figure}

\noindent \textbf{Qualitative analysis.} 
We visualize the result of node sparsification in Fig. \ref{figure:visualize_sparsification}. Given $0 < q_{v} < 1$, \ours\ learns to drop visual tokens by retaining salient patches and ignoring ambient patches. 
\newline

\noindent \textbf{GPU memory usage.} Peak memory usage for configuration $(K_{l}, K_{r}, G) = (1, 3, 56)$ was 8.76GB per GPU and $(K_{l}, K_{r}, G) = (1, 5, 56)$ was 9.61GB per GPU. This included a copy of the model, current batch, and gradients.

%% file: sec/5_conclusion.tex
\section{Conclusions}
\label{sec:conclusion}

We introduce \ours, a sparse video-text architecture equipped with multi-frame reasoning capability for egocentric video understanding. \ours\ employs two forms of sparsity: edge sparsity that limits the query-key communications between tokens in self-attention, and node sparsity that discards uninformative visual tokens. \ours\ outperforms dense transformer baselines on the EgoMCQ task with significantly lower peak memory and compute requirements thanks to the sparsity. This shows that sparse architecture such as \ours\ is a potential foundation model choice especially for pretraining on memory-bound devices.